\def\year{2022}\relax
\documentclass[letterpaper]{article} 
\usepackage{aaai22}  
\usepackage{times}  
\usepackage{helvet}  
\usepackage{courier}  
\usepackage[hyphens]{url}  
\usepackage{graphicx} 
\usepackage{natbib}  
\usepackage{caption} 
\DeclareCaptionStyle{ruled}{labelfont=normalfont,labelsep=colon,strut=off} 
\usepackage{algorithm}
\usepackage{algorithmic}

\usepackage{newfloat}
\usepackage{listings}
\usepackage{amstext}
\floatstyle{ruled}
\newfloat{listing}{tb}{lst}{}
\floatname{listing}{Listing}
\title{Accelerating Understanding of Scientific Experiments \\with End to End Symbolic Regression}
 \author {
     Nikos Ar\'echiga,
     Francine Chen,
     Yan-Ying Chen,\\
     Yanxia Zhang,
     Rumen Iliev,
     Heishiro Toyoda,
     Kent Lyons
 }
\affiliations{
    Toyota Research Institute\\
    Los Altos, CA\\
    \{nikos.arechiga,
    francine.chen,
    yan-ying.chen,
    yanxia.zhang, 
    rumen.iliev,
    heishiro.toyoda,
    kent.lyons\}@tri.global
}

\begin{document}

\maketitle

\begin{abstract}
We consider the problem of learning free-form symbolic expressions from raw data, such as that produced by an experiment in any scientific domain.
Accurate and interpretable models of scientific phenomena are the cornerstone of scientific research. Simple yet interpretable models, such as linear or logistic regression and decision trees often lack predictive accuracy. Alternatively, accurate blackbox models such as deep neural networks provide high predictive accuracy, but do not readily admit human understanding in a way that would enrich the scientific theory of the phenomenon. Many great breakthroughs in science revolve around the development of parsimonious equational models with high predictive accuracy, such as Newton's laws, universal gravitation, and Maxwell's equations. Previous work on automating the search of equational models from data combine domain-specific heuristics as well as computationally expensive techniques, such as genetic programming and Monte-Carlo search. We develop a deep neural network (MACSYMA) to address the symbolic regression problem as an end-to-end supervised learning problem. MACSYMA can generate symbolic expressions that describe a dataset. The computational complexity of the task is reduced to the feedforward computation of a neural network. We train our neural network on a synthetic dataset consisting of data tables of varying length and varying levels of noise, for which the neural network must learn to produce the correct symbolic expression token by token. Finally, we validate our technique by running on a public dataset from behavioral science.
\end{abstract}

\section{Introduction}
Accurate and interpretable models of real-world phenomena are the cornerstone of scientific progress. An accurate model makes high-quality predictions of the phenomenon under study. The recent rise of deep learning yields a general methodology for the production of high-accuracy prediction models: fit a deep neural network that accurately predicts the dependent variables of an experiment given values of the independent values. A key drawback of this approach, however, is that the deep neural network is not interpretable, in the sense that it is not feasible for a scientist to inspect the network and compose its model with other models from scientific theory.

Instead, we would like a model that can provide a parsimonious equational form that models the data under consideration. The general problem of inferring symbolic equations from data is known as \emph{symbolic regression}.
The general problem of symbolic regression is computationally difficult because equations are strings of symbols, and the number of possible strings grows exponentially with the length of the string. A direct attempt to enumerate and test all possible equational strings would take longer than the age of the universe to find the equation of interest \cite{udrescuAIFeynmanPhysicsinspired2020}.

There is a rich of body of work around symbolic regression with metaheuristics such as 
sparse regression\cite{bruntonDiscoveringGoverningEquations2016, quadeSparseIdentificationNonlinear2018}
and genetic programming
\cite{searsonGPTIPSOpenSource2010, pitzerSmoothSymbolicRegression2021},
including the commercial software tool Eureqa \cite{dubcakovaEureqaSoftwareReview2011}. Such metaheuristics, however, are computationally expensive.

In this work, we develop a neural network, called MACSYMA, to address the symbolic regression problem as an end-to-end supervised learning problem. We use supervised learning to map directly from experimental data tables to symbolic equations that describe that data. The computational complexity of our approach is reduced to the complexity of doing a feedforward pass of a neural network. Our training set accounts for variable length as well as noisy datasets. Our existing model has an output of maximum length, but in future work we will generalize our approach to use a recurrent neural network that is capable of producing sequences of any length. We demonstrate our approach on a public dataset from behavioral science that illustrates prospect theory \cite{kahnemanProspectTheoryAnalysis1979}.

\section{Related Work}
The problem of synthesizing symbolic expressions from raw data, commonly referred to as symbolic regression, has a rich history within the research community. Established approaches rely on the use of computationally expensive genetic algorithms. In this section, we review some recent work on symbolic regression.

The work in \cite{udrescuAIFeynmanPhysicsinspired2020} describes a physics-inspired method for symbolic regression, in which data is analyzed by a multistage process involving heuristics such as dimensional analysis, searching for symmetries, and others. In contrast, our work does not rely on manually encoded heuristics or domain-specific transformations, but instead directly transforms from raw data to a symbolic form.

In \cite{itenDiscoveringPhysicalConcepts2020}, the authors develop a neural network approach to modeling dynamic equations, predicting the evolution of dynamical systems over time. Their approach is similar to ours in the use of neural networks, but distinct in that we do not limit our attention to dynamical systems, but merely provide data of given variables and request that the model extract symbolic relationships.

\cite{cranmerDiscoveringSymbolicModels2020a} uses neural networks with a graph structure, intended to model the interactions between particles and forces, and uses a genetic programming package to extract the symbolic model. The usage of graph neural networks is intended to bias the model to produce symbolic expressions that are amenable to model sets of particles subject to forces, such as gravity or electromagnetism. In contrast, our approach completely elides the need for computationally expensive genetic programming techniques and is not limited to application domains with force-particle interactions.

In \cite{kusnerGrammarVariationalAutoencoder2017}, the authors train a variational autoencoder (VAE) over strings of equations. The VAE consists of an encoder network and a decoder network. The encoder network encodes equation strings into a latent space, which the decoder learns to reconstruct into equation strings. After training, the task of symbolic regression is performed by optimizing over the latent space to find a point that decodes to an equation that minimizes error over the given data. This optimization is computationally expensive.

An alternative approach in \cite{brenceProbabilisticGrammarsEquation2021} models equations by a context-free grammar. Each production rule of the grammar is given a probability, and equations are produced through Monte Carlo search. 

Both the latent-space optimization of \cite{kusnerGrammarVariationalAutoencoder2017} and the Monte Carlo search of \cite{brenceProbabilisticGrammarsEquation2021} are computationally expensive. In contrast, our approach performs inference by a feedforward pass of a neural network.


\section{Methodology}

\begin{figure}[ht]
    \centering
    \includegraphics[width=0.475\textwidth]{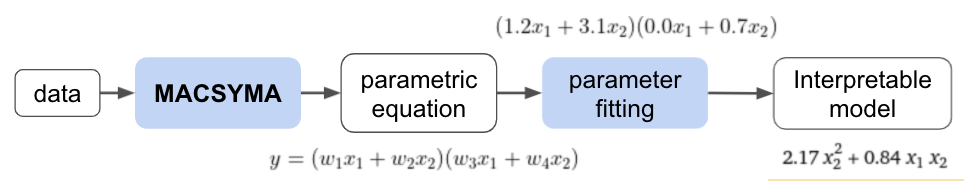}
    \caption{Usage flowchart. From a (possibly noisy) raw data table, MACSYMA produces a symbolic equation with unknown parameters $w_i$ (in this case a quadratic form). A downstream regression task fits the parameters, and the equation can be simplified by off-the-shelf algebra software.}
    \label{fig:usage}
\end{figure}

The usage of our system is illustrated in Figure \ref{fig:usage}. First, a data table is collected from an experiment. This data table has columns corresponding to the independent variables as well as the dependent variable. For example, suppose an experiment contains the independent variables \(x_1\) and \(x_2\), and the goal is to obtain an equational model for the effect of these variables on the dependent variable \(y\). Then, the data table consists of observations of \(x_1\), \(x_2\), and \(y\). Next, this data table is fed into our neural network MACSYMA, which outputs an equation, for example \(y = (w_1 x_1 + w_2 x_2) (w_3 x_1 + w_4 x_2)\), where \(\{w_1, w_2, w_3, w_4\}\) are parameters of the equation. The final step is to perform a conventional regression procedure to fit values for the parameters \(w_i\), resulting in an equational model of the data table.

\subsection{Synthetic dataset}
To train the model, we developed a dataset that consists of noisy tabular data and the symbolic expression that corresponds to it. We generated this dataset by first collecting a series of symbolic expressions, including linear and polynomial equations, the min, max, argmin and argmax functions, and transcendental functions, such as exponentials, logarithms, sigmoids, sines, and cosines. In future work, we will add equations from the public dataset in \cite{udrescuAIFeynmanPhysicsinspired2020}.

Each symbolic expression contains parameters and variables. The variables represent placeholders for the data to be modeled, and the parameters represent values to be fitted by the downstream regression task.
\begin{figure}[ht]
    \centering
    \includegraphics[width=0.475\textwidth]{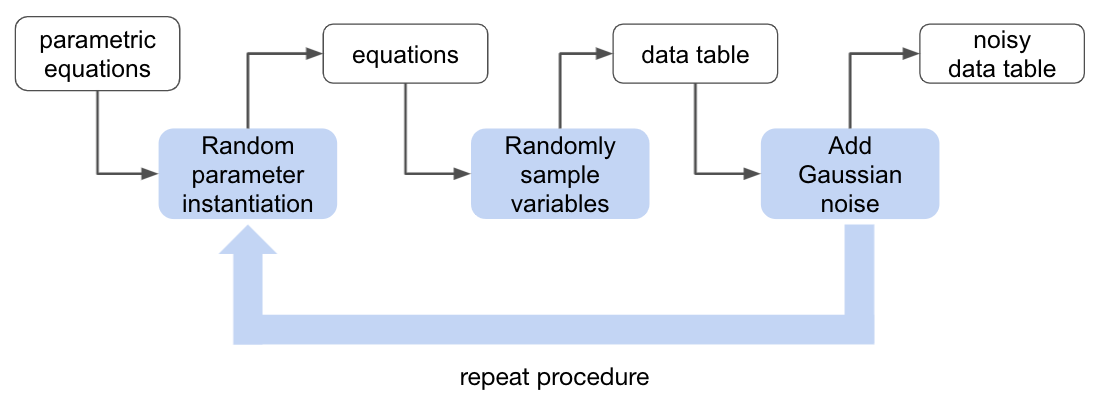}
    \caption{Data generation flowchart}
    \label{fig:datagen_flowchart}
\end{figure}
Figure \ref{fig:datagen_flowchart} illustrates the process for generating the dataset. We begin with a corpus of parametric equations. Each equation contains placeholders for numeric parameters as well as placeholders for data variables. Next, we randomly select values for the parameters of the equations. Then, we generate a data table for the equation by repeatedly plugging random values into the data variables and evaluating the equation. We allow for data tables of varying length, but not to exceed a total of $200$ numerical values, including all dependent variable values as well as dependent variable values. Finally, we add Gaussian noise to the values of the dependent variables, to model the effect of measurement noise. For each equation, this process is repeated with ten different noise levels and with different randomly selected parameter instantiations to complete 5000 pairings between noisy data tables and their corresponding equations. By using different parameter instantiations, we encourage MACSYMA to learn the general mapping between data and its symbolic structure, generalizing across parameter values.


\subsection{Neural network architecture}
\begin{figure}[ht]
    \centering
    \includegraphics[width=0.475\textwidth]{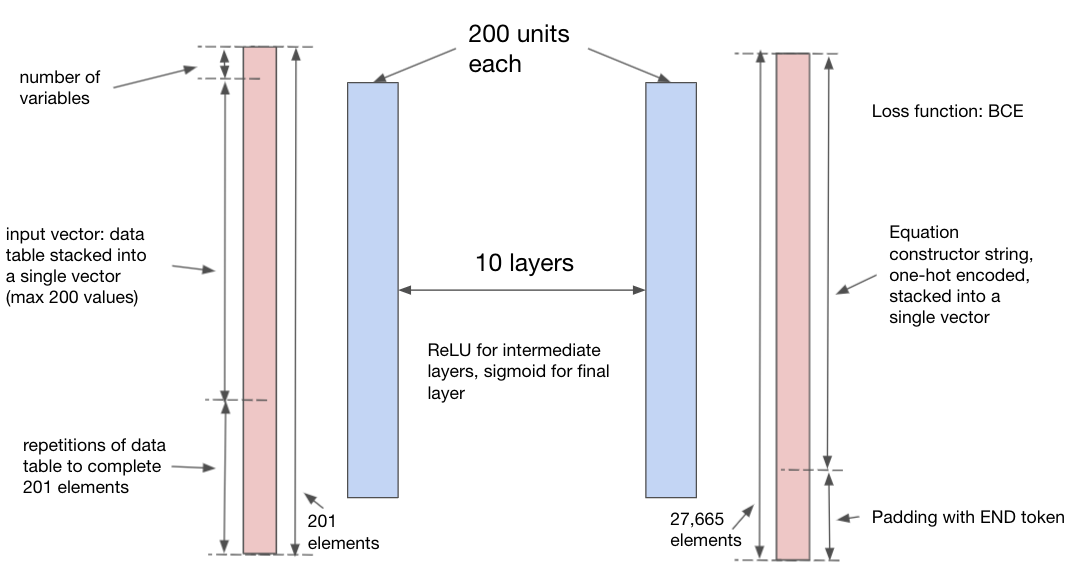}
    \caption{Neural network architecture}
    \label{fig:nn_architecture}
\end{figure}

Figure \ref{fig:nn_architecture} diagrams the neural network architecture. The input vector consists of 201 elements. The first element is the number of data variables that appear in the equation, and the remainder of the vector is obtained by stacking the columns of the data table. Since the data tables have variable length, the stacked columns are repeated multiple times and then truncated to yield a vector of length $201$. The network has 10 layers with 200 units each, and the output layer has $27,655$ elements, which is the number of elements needed to represent the longest equation in our corpus. The nonlinearity at all layers are ReLUs except for the last layer, which has a sigmoid nonlinearity. The outputs are interpreted as binary bits, which are decoded as one-hot encodings of the tokens in an equation. The loss function used is binary cross-entropy.

\section{Model training}
\begin{figure}[ht]
    \centering
    \includegraphics[width=0.4\textwidth]{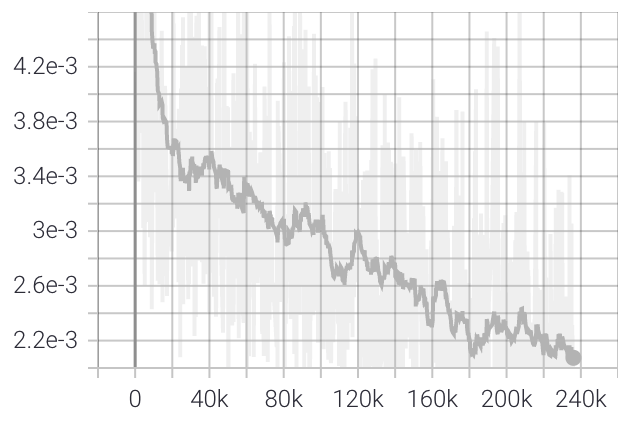}
    \caption{Training curve at each batch.}
    \label{fig:training_curve}
\end{figure}

\begin{figure}[ht]
    \centering
    \includegraphics[width=0.4\textwidth]{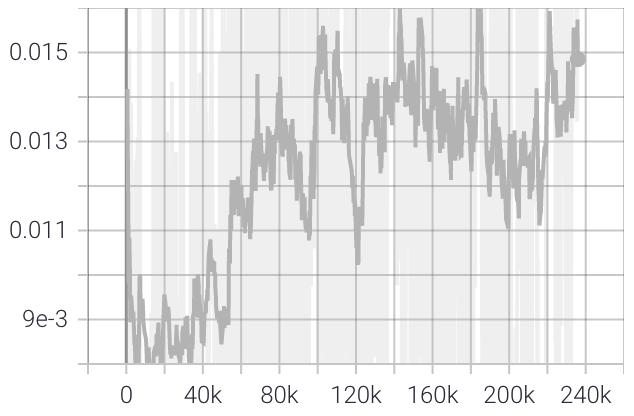}
    \caption{Validation curve. We continued training past the point where the validation loss was increasing because the percentage of parsable formulas continued to increase. We will investigate the trade-off between validation loss and percentage of parsable formulas in future work}
    \label{fig:xval_curve}
\end{figure}

We trained for $100$ epochs using a batch size of $64$ training stimuli. Each training stimulus consists of an input data table and an output equation. Figure \ref{fig:training_curve} shows the training loss, and Figure \ref{fig:xval_curve} shows the cross-validation loss. We continued training past the point where cross-validation error was increasing because the percentage of formulas that parsed correctly increased past the point at which cross-validation loss was minimal. In future work, we will investigate the trade-offs between these two metrics, and their effect on the performance of the equations generated.

An important challenge in the use of neural network models for symbolic regression is whether the model will efficiently learn the grammar of logical formulas. Previous work sidesteps this issue, for example in the use of a grammar-based checker which selects the most likely token that produces a grammatically correct formula, and backtracks when no such token exists \cite{kusnerGrammarVariationalAutoencoder2017}. Other approaches, such as \cite{brenceProbabilisticGrammarsEquation2021}, make the grammar of the logical formulas the basic scaffolding of the generative model, and only allow learning probabilities with which each production rule is applied. Approaches that rely on genetic programming to construct the final formula, such as \cite{dubcakovaEureqaSoftwareReview2011} (and also \cite{cranmerDiscoveringSymbolicModels2020a}, since the formula is extracted from the graph neural network by a genetic program) can constrain the search space by only considering solutions that parse correctly.

\begin{figure}[ht]
\centering
    \includegraphics[width=0.4\textwidth]{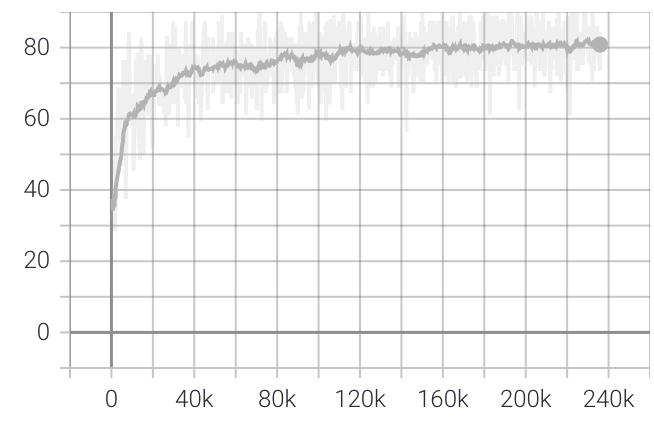}
    \caption{Percentage of formulas that parse correctly at each batch.}
    \label{fig:grammar_learning}
\end{figure}

Our approach, however, does not strictly enforce the grammar of logical formulas, and hence we rely on the ability of the neural network to learn the grammar. To evaluate the ability of the network to learn the grammar of logical formulas, Figure \ref{fig:grammar_learning} demonstrates a plot of the network's ability to produce formulas that parse correctly. This is done by attempting to parse the formulas produced every $100$ training steps in a batch of $64$. The y-axis represents the percentage of formulas that parsed correctly. The percentage of formulas that parse correctly seems to plateau around $80\%$\. To the best of our knowledge, we are the first group to evaluate the ability of a neural network without grammatical scaffolding to produce equations that parse correctly. In the future, we will investigate the use of beam search, both to increase the percentage of formulas that parse correctly and to allow a human to interactively inspect several candidate formulas. The human interactivity could enable a scientist to select formulas with domain-specific meaning.

\section{Results on an unseen dataset}
As a case study, we use a subset of the Social Expected Utility (SEU) dataset \cite{ilievSocialExpectedUtility2021}. This dataset models a series of gambles presented to a survey participant. Each gamble corresponds of two possibilities, the first with a possibility of attaining a dollar value $V_1$ with probability $p_1$ versus a possibility of attaining dollar value $V_2$ with probability $p_2$.

The SEU dataset contains variations on this survey type, including ones in which the participant will benefit from one gamble, and the next survey participant will benefit from the second gamble. In this case study, we focus only on the case when both gambles are for the survey participant, and also both possible values are non-negative, $V_1 \ge 0$ and $V_2 \ge 0$.

One way to model this type of choice is to represent the \emph{subjective value} (SV) of each alternative $(p_i, V_i)$ by Equation \ref{eqn:subjective_value} \cite{yuSteeperDiscountingDelayed2017},
\begin{equation}\label{eqn:subjective_value}
    SV_i = p_i V_i^\alpha
\end{equation}
where $\alpha$ varies from person to person, and is commonly interpreted as a \emph{risk aversion parameter}. (Higher $\alpha$ is interpreted as a larger risk tolerance, \cite{yuSteeperDiscountingDelayed2017}). A direct way to model the choice of alternatives is to simply select the choice with the larger subjective value, 
\begin{equation}\label{eqn:sv_choice_max}
\text{choice} = \text{arg}\max(SV_1, SV_2)
\end{equation}
An alternative \cite{yuSteeperDiscountingDelayed2017} is to add an additional ``noise'' parameter, $\beta$, which scales a sigmoid. Equation \ref{eqn:sv_choice_sigmoid} yields a number between $1$ and $2$, and may be rounded to give a prediction of the choice the participant will make.
\begin{equation}\label{eqn:sv_choice_sigmoid}
    \text{choice} = \sigma(\beta(SV_1 - SV_2)) + 1
\end{equation}
The structure of both Equation \ref{eqn:sv_choice_max} and Equation \ref{eqn:sv_choice_sigmoid} appear in the training corpus of MACSYMA. The SEU dataset is used for testing only. When we run MACSYMA on the SEU dataset, we recover Equation \ref{eqn:sv_choice_max}, which is one of the well-known theory-based models.

\section{Discussion and future work}
An important drawback of our existing neural network architecture is that there is a maximum size to the formula that can be produced. Furthermore, using a static, non-recurrent architecture means that increasing the maximum formula length means that the size of the neural network will increase exponentially, because all of the layers are fully connected. In future work, we will explore the use of a recurrent architecture to address these issues. Additionally, we will benchmark our approach against the public AI Feynman dataset \cite{udrescuAIFeynmanPhysicsinspired2020}, extending the range of equations that are supported and enabling comparison against other approaches. 
We will also implement beam search at the output of the neural network to enable a human operator to interactively search for formulas that may have a specific interesting structure to the domain.

\bibliography{bibliography}

\begin{thebibliography}{13}
\providecommand{\natexlab}[1]{#1}

\bibitem[{Brence, Todorovski, and D{\v
  z}eroski(2021)}]{brenceProbabilisticGrammarsEquation2021}
Brence, J.; Todorovski, L.; and D{\v z}eroski, S. 2021.
\newblock Probabilistic {{Grammars}} for {{Equation Discovery}}.
\newblock \emph{Knowledge-Based Systems}, 224: 107077.

\bibitem[{Brunton, Proctor, and
  Kutz(2016)}]{bruntonDiscoveringGoverningEquations2016}
Brunton, S.~L.; Proctor, J.~L.; and Kutz, J.~N. 2016.
\newblock Discovering Governing Equations from Data by Sparse Identification of
  Nonlinear Dynamical Systems.
\newblock \emph{Proceedings of the National Academy of Sciences}, 113(15):
  3932--3937.

\bibitem[{Cranmer et~al.(2020)Cranmer, {Sanchez-Gonzalez}, Battaglia, Xu,
  Cranmer, Spergel, and Ho}]{cranmerDiscoveringSymbolicModels2020a}
Cranmer, M.; {Sanchez-Gonzalez}, A.; Battaglia, P.; Xu, R.; Cranmer, K.;
  Spergel, D.; and Ho, S. 2020.
\newblock Discovering {{Symbolic Models}} from {{Deep Learning}} with
  {{Inductive Biases}}.
\newblock \emph{arXiv:2006.11287 [astro-ph, physics:physics, stat]}.

\bibitem[{Dub{\v c}{\'a}kov{\'a}(2011)}]{dubcakovaEureqaSoftwareReview2011}
Dub{\v c}{\'a}kov{\'a}, R. 2011.
\newblock Eureqa: Software Review.
\newblock \emph{Genetic Programming and Evolvable Machines}, 12(2): 173--178.

\bibitem[{Iliev(2021)}]{ilievSocialExpectedUtility2021}
Iliev, R. 2021.
\newblock Social {{Expected Utility Dataset}}.
\newblock https://github.com/TRI-MAC/SEU.

\bibitem[{Iten et~al.(2020)Iten, Metger, Wilming, {del Rio}, and
  Renner}]{itenDiscoveringPhysicalConcepts2020}
Iten, R.; Metger, T.; Wilming, H.; {del Rio}, L.; and Renner, R. 2020.
\newblock Discovering Physical Concepts with Neural Networks.
\newblock \emph{Physical Review Letters}, 124(1): 010508.

\bibitem[{Kahneman and Tversky(1979)}]{kahnemanProspectTheoryAnalysis1979}
Kahneman, D.; and Tversky, A. 1979.
\newblock Prospect {{Theory}}: An {{Analysis}} of {{Decision}} under {{Risk}}.
\newblock \emph{Econometrica}, 47(2): 263.

\bibitem[{Kusner, Paige, and
  {Hern{\'a}ndez-Lobato}(2017)}]{kusnerGrammarVariationalAutoencoder2017}
Kusner, M.~J.; Paige, B.; and {Hern{\'a}ndez-Lobato}, J.~M. 2017.
\newblock Grammar {{Variational Autoencoder}}.
\newblock \emph{arXiv:1703.01925 [stat]}.

\bibitem[{Pitzer and Kronberger(2021)}]{pitzerSmoothSymbolicRegression2021}
Pitzer, E.; and Kronberger, G. 2021.
\newblock Smooth {{Symbolic Regression}}: Transformation of {{Symbolic
  Regression}} into a {{Real}}-Valued {{Optimization Problem}}.
\newblock \emph{arXiv:2108.03274 [cs]}, 9520: 375--383.

\bibitem[{Quade et~al.(2018)Quade, Abel, Kutz, and
  Brunton}]{quadeSparseIdentificationNonlinear2018}
Quade, M.; Abel, M.; Kutz, J.~N.; and Brunton, S.~L. 2018.
\newblock Sparse {{Identification}} of {{Nonlinear Dynamics}} for {{Rapid Model
  Recovery}}.
\newblock \emph{Chaos: An Interdisciplinary Journal of Nonlinear Science},
  28(6): 063116.

\bibitem[{Searson, Leahy, and Willis(2010)}]{searsonGPTIPSOpenSource2010}
Searson, D.~P.; Leahy, D.~E.; and Willis, M.~J. 2010.
\newblock {{GPTIPS}}:An {{Open Source Genetic Programming Toolbox For Multigene
  Symbolic Regression}}.
\newblock \emph{Hong Kong}, 4.

\bibitem[{Udrescu and Tegmark(2020)}]{udrescuAIFeynmanPhysicsinspired2020}
Udrescu, S.-M.; and Tegmark, M. 2020.
\newblock {{AI Feynman}}: A Physics-Inspired Method for Symbolic Regression.
\newblock \emph{Science Advances}, 6(16): eaay2631.

\bibitem[{Yu et~al.(2017)Yu, Lee, Katchmar, Satterthwaite, Kable, and
  Wolf}]{yuSteeperDiscountingDelayed2017}
Yu, L.~Q.; Lee, S.; Katchmar, N.; Satterthwaite, T.~D.; Kable, J.~W.; and Wolf,
  D.~H. 2017.
\newblock Steeper Discounting of Delayed Rewards in Schizophrenia but Not
  First-Degree Relatives.
\newblock \emph{Psychiatry Research}, 252: 303--309.

\end{thebibliography}

\end{document}